\documentclass[10pt,journal,compsoc]{IEEEtran}
\ifCLASSOPTIONcompsoc
  \usepackage[nocompress]{cite}
\else
  \usepackage{cite}
\fi
\ifCLASSINFOpdf
 \usepackage[pdftex]{graphicx}
 \graphicspath {{Images/}}
\else
\fi
\usepackage{amsmath,amssymb}
\usepackage{array}

\ifCLASSOPTIONcompsoc
	\usepackage[caption=false,font=footnotesize,labelfont=sf,textfont=sf]{subfig}
\else
  \usepackage[caption=false,font=footnotesize]{subfig}
\fi
\makeatletter
\long\def\@IEEEtitleabstractindextextbox#1{\parbox{0.922\textwidth}{#1}}
\makeatother
\usepackage{hyperref}

\begin{document}
%
\title{Wi2Vi: Generating Video Frames from WiFi CSI Samples}
%
%
%
%

\author{Mohammad~Hadi~Kefayati,
        ~Vahid~Pourahmadi,
        ~and~Hassan~Aghaeinia,
\IEEEcompsocitemizethanks{\IEEEcompsocthanksitem All authors are with the Department
of Electrical Engineering, Amirkabir University of Technology (Tehran Polytechnic), Tehran,
Iran.\protect\\
Corresponding author is Vahid Pourahmadi,\hfil\break E-mail: v.pourahmadi@aut.ac.ir }}
\IEEEtitleabstractindextext{%
\begin{abstract}
Objects in an environment affect electromagnetic waves. While this effect varies across frequencies, there exists a correlation between them, and a model with enough capacity can capture this correlation between the measurements in different frequencies. In this paper, we propose the Wi2Vi model for associating variations in the WiFi channel state information with video frames.  The proposed Wi2Vi system can generate video frames entirely using CSI measurements. The produced video frames by the Wi2Vi provide auxiliary information to the conventional surveillance system in critical circumstances. Our implementation of the Wi2Vi system confirms the feasibility of constructing a system capable of deriving the correlations between measurements in different frequency spectrums.
 
\end{abstract}

}

\maketitle

\IEEEdisplaynontitleabstractindextext

%
\IEEEpeerreviewmaketitle


\section{Introduction}\label{sec:introduction}

Electromagnetic (EM) waves carry out most information we sense in this world. We perceive our surroundings using their visual range, i.e., visible light. Wireless communications use lower frequency ranges of the EM waves spectrum, i.e., radio frequencies (RF). We sense visible light using our eyes, and we have created communication systems to sense and use radio frequencies.

We recognize the existence of objects in the environment by their effect on EM waves, like the reflection of light and other EM waves off an object. Even though the effect of an object on EM waves varies as the frequency of the EM waves varies, there is a correlation between the measurements in different frequencies. Specifically, the measurements in RF ranges and optical range are correlated as they represent the existence of the same objects in the environment. Due to the complex relationship between the measurements in different frequencies, there is no mathematical scheme that can explain this duality. However, a model with enough capability can infer an association between these complex changes, resulting in mapping from one domain to another. 

Application of the similarity between the visible and RF spectrum has been used in the literature before, for example, ray tracing 
\cite{anderson1993ray} 
used in RF propagation modeling has its origins in optics \cite{glassner1989introduction}, and the similarities between EM waves in RF and visible spectrum has made it reasonable to model the wireless channel using ray tracing models. 

\begin{figure}[!t]
\centering
\includegraphics[width=.9\columnwidth]{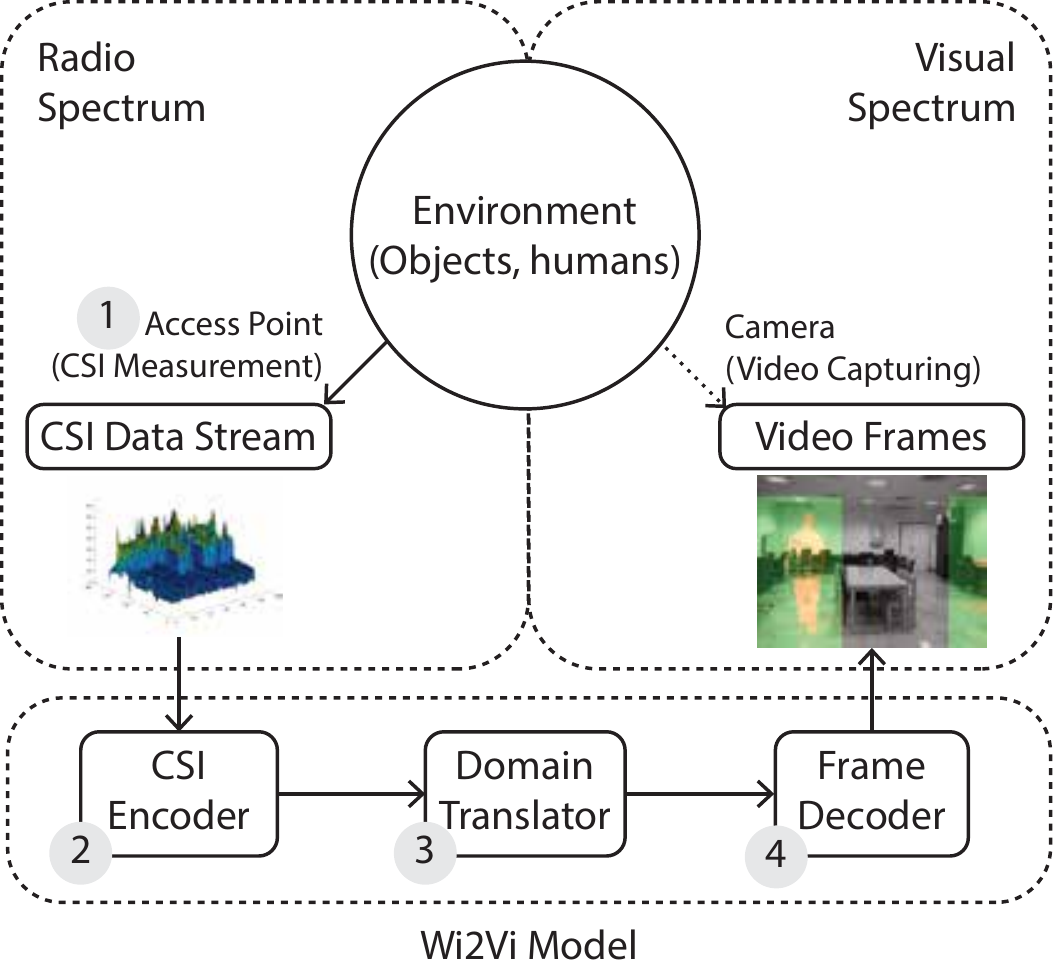}
\caption{The Wi2Vi theory of operation}
\label{fig:CSI-env-Im-model}
\end{figure}

A method to translate information from other frequency ranges of the EM spectrum to our visual range can extend our comprehension of the world. Complete translation from one of part of the EM spectrum to another is a complex task since various materials affect EM waves in separate frequencies differently. Two substances can have a similar effect on EM waves in one frequency range and have a widely different effect in another. This is the main reason that finding even a partial relationship between the two bands is very challenging though precious. 

To overcome this inherent complexity, in this paper, we present a machine learning model for translating the perceived data in the WiFi spectrum to visual data, see figure \ref{fig:CSI-env-Im-model}. Our model uses the CSI data from the commercially off the shelf (COTS) WiFi access points (APs) as a measurement of the environment in the WiFi frequency spectrum (step 1 in figure \ref{fig:CSI-env-Im-model}). The \textit{CSI Encoder} then extracts information in the latent RF domain (step 2 in figure \ref{fig:CSI-env-Im-model}). The \textit{Domain Translator} associates the latent RF domain to the latent visual domain (step 3 in figure \ref{fig:CSI-env-Im-model}). The \textit{Frame Decoder} finally generates images with details on the static background and dynamic objects like the human subject in the environment from the information in the latent visual domain (step 4 in figure \ref{fig:CSI-env-Im-model}). Running this procedure on a time series of CSI data, Wi2Vi (WiFi-to-Video) can generate video frames as if a camera from the scene has captured them.

As far as we know, the proposed Wi2Vi model is the first model able to create a video sequence of the human subject and the background objects in the environment entirely from the WiFi CSI sample sequence. The Wi2Vi model partially extracts the underlying correlation between CSI measurements with the scene's video frames and uses this to generate video frames for critical situations when the actual video frames are not accessible.

The Wi2Vi model uses the ration explained in the beginning (figure \ref{fig:CSI-env-Im-model}), and is implemented as a deep neural network in three parts. Firstly, The model extracts information about the object (human occupant) using the initial \textit{CSI Encoder} layers. After this, the translation layers render this information from the WiFi spectrum's latent domain to the visual spectrum's latent domain. Lastly, the \textit{Frame Decoder} layers generate images from the given underlying representation in the visual spectrum.

We have also developed new unique methods in the data preprocessing, to deal with the limits in CSI data obtained from COTS devices, making it possible to generate video sequences from CSI sequences from a single receiver AP with only three antennas. Unlike most previous studies, our approach does not require any specific hardware calibration procedure.

Previous studies in WiFi imaging, e.g., \cite{zi2019wi} use modified WiFi setups with multiple APs and provide limited results using radar-based approaches. Apart from their limitations and costs in implementations, their results are limited since they ignore the underlying translation in EM wave spectrums. Our deep model extracts this translation using the camera output in the initial training phase. The execution phase of the Wi2Vi model only requires CSI samples, massively limiting its deployment cost. 

Reference \cite{wang2019person} has recently developed a model to percept humans using WiFi CSI. Their work is an augmentation of their previous studies in human pose estimation using WiFi measurements \cite{wang2019can,wang2018csi}. They focus on the human perception problem, estimating pose, and the related parameters in human perception, however, we are focusing on translating data from RF measurements (CSI data) into the visual spectrum. We have taken the occupant's sensing as a demonstration for our work since it has fashionable practical applications. 
Even though our evaluations for the Wi2Vi model is more focused on the human occupant, our model can be developed and used to translate CSI measurements in other environments with limited dynamics to the visual domain. 

Prior work related to our problem, i.e., creating video sequences using WiFi signals are mainly categorized under WiFi imaging. WiFi imaging techniques fall into two categories: Ghost imaging using scattering models and radar-theory based imaging. Even though our objective is related to them, our approach to model the relationship is entirely different, and Wi2Vi is the first model able to deliver video from the CSI data stream. Machine learning models were previously used in wireless sensing (WS) field for tasks like localization, gesture and activity recognition, and pose estimation. Our main contributions in this work are as follows

\begin{itemize}
	\item Proposing a new approach to the WiFi imaging problem, RF spectrum to visual spectrum mapping via machine learning
	
	\item Implementing Wi2Vi system using commercial off the shelf (COTS) devices, creating video frames from WiFi CSI measurements
	\item Introducing a new approach to sync video and camera data, a requirement for generating video frames from the CSI data. 
	\item Proposing a new data augmentation method based on the dropout \cite{srivastava2014dropout}, which we named "dropin"
\end{itemize}

\begin{figure*}[!t]
\centering
\subfloat[Normal conditions (Wi2Vi is being trained)]{\includegraphics[width=.97\columnwidth]{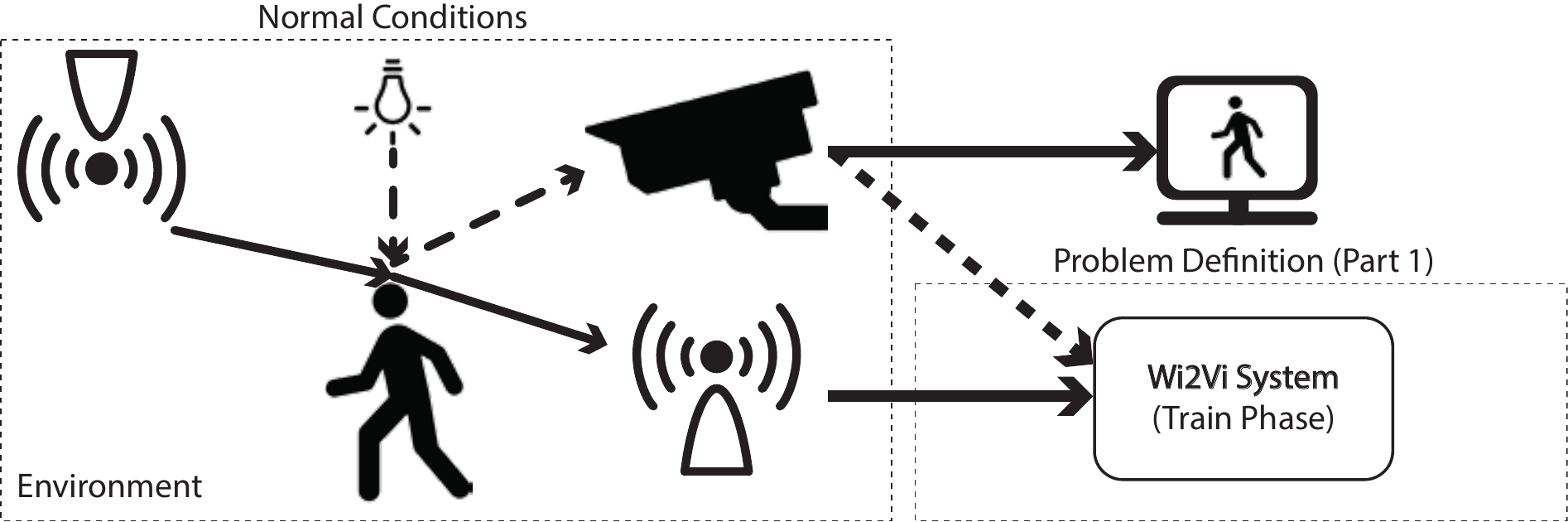}%
\label{fig:Normal-Scenario}}
\hfil
\subfloat[Critical conditions (Wi2Vi is being exploited)]{\includegraphics[width=.97\columnwidth]{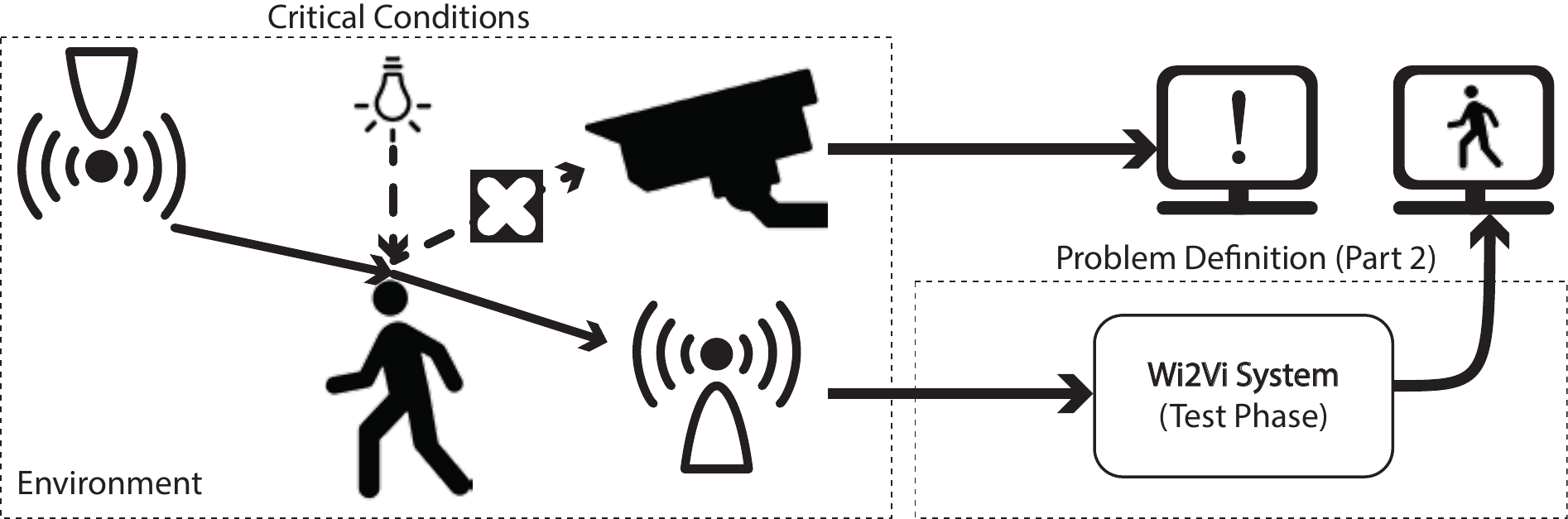}%
\label{fig:Critical-Scenario}}
\caption{Practical surveillance problem definition utilizing the Wi2Vi model}
\label{fig:problem-scenarios}
\end{figure*}
 
\section{Related Works}\label{sec:Related-Works}
In this section we briefly review previous studies related to our problem.
\subsection{WiFi Sensing}
WiFi is the most common wireless technology used in almost all indoor environments. Hence using it to sense comes at no cost. Wireless sensing (WS) dates back to wireless sensor networks (WSN) studies. Similar to WSN, some early studies in WiFi sensing use WiFi received signal strength indicator (RSSI) signals for probing the environment, and for sensing.  OFDM and MIMO technologies adopted in 802.11n/ac provide better channel monitoring capabilities at the WiFi receiver side, and open source projects, e.g., CSI tool \cite{Halperin_csitool} and Atheros CSI tool \cite{Xie:2015:PPD:2789168.2790124-} provide researchers with convenient access methods to capture CSI data. These projects have enabled researchers to investigate wireless sensing further using WiFi CSI. Ever since then, researchers applied WS to different problems. 

Several comprehensive surveys on WS in general, e.g, \cite{ma2019wifi,al2019channel} and other surveys on the applications of WS, e.g., 
\cite{yousefi2017survey,jiang2018smart,wang2019survey} 
have reviewed recent trends of this field in details. 


\subsubsection{Pose estimation}
The escalating augmented reality (AR) era requires estimating human posture and gestures. Among the device free approaches for satisfying this requirement, vision-based systems are the most adopted scheme \cite{zhang2019comprehensive}. But they face two issues: 1) \emph{Specific lighting requirements} 2) \emph{Privacy violation}. Low light and high contrast scenes degrade the accuracy of the pose estimation in such systems. A third party attacker may compromise the security of the system, and data leak from the installed cameras in households can result in serious violation of the user's privacy. WS can provide an acceptable solution to the pose estimation without facing these issues.

The recently proposed "Person in WiFi" model \cite{wang2019person} 
generates joint heat maps and part affinity fields from WiFi signals. These parameters express the human pose for pose estimation purposes.
The neural network model they use first upsamples the CSI samples, then employs multiple Resnet 
and UNet structures to regress the pose parameters. The 2D body segmentation mask they reconstruct is a binary-valued image of the person. Hence, this mask has similarities to the results we present for the first use case of the Wi2Vi model. Our goal and modeling structure are entirely different from \cite{wang2019person}. Moreover, the second use case that we present differentiates our objective and emphasize on capabilities of the Wi2Vi model.

\subsection{WiFi Imaging}
The Wision \cite{huang2014feasibility} approach is an early study in WiFi imaging and has used the 2D Fourier transform to discriminate the arriving rays in different directions. After this pioneering study on the application of WiFi signals for imaging purposes, two major approaches for imaging using WiFi signals are initiated.

The first approach use radar principles and apply them to CSI data for generating images \cite{zhou2015sensorless}. Radar-based approaches focus on differentiating between wireless signal reflection paths and extracting the rays bouncing off the human body. The multiple signal classification (MUSIC) method is the primary choice in these scehemes for distinguishing reflecting paths \cite{zi2019wi,adib2013see}. Since the reflected path from the human body is the reflection from the transmitter signal, i.e., coherent to the transmitted signal,  the smooth MUSIC is a more logical choice \cite{kotaru2015spotfi}.

The second approach to use WiFi transmissions for creating an image is to use scattering principles and microwave imaging theorems for modeling the imaging problem 
\cite{holl2017holography}. 
 
Reference \cite{holl2017holography} used the holography principles on raw signal measurements in the WiFi band, i.e., non-CSI data. They acquired measurements in a phase-coherent fashion by two antennas, a fixed reference antenna, and a scanning antenna moving in a 2D plane. They were able to produce holograms of metal objects using these measurements.   

Reference \cite{luo2018wi} and \cite{wang2016microwave} use microwave ghost imaging principles in distributed antenna settings to approach the problem. WiFi transmissions include a known preamble part, which helps in extracting CSI and data recovery, and a random data payload part. Unlike radar-based approaches were the CSI part of WiFi transmissions is the data source for developing images, in ghost imaging, incoherence and randomness in the data part of WiFi transmissions are favored and utilized. The main weakness of ghost imaging approaches is the numerous signal acquisition points required.

Even though reference \cite{zhang2019feasibility}'s objective was not imaging, it has recently considered the possibility of detecting the object's location and type using WiFi signals. Their IntuWition model captures the polarization of the reflecting waves from objects in an environment and enables them to demonstrate that it is possible to infer material properties.

\section{Wi2Vi overview\label{sec:wi2vi-overview}}
As mentioned previously in section \ref{sec:introduction}, the human occupant imaging is selected in our use cases of the Wi2Vi model. In this practical scenario, Wi2Vi acts as an additional modality to augment the visual perception of the environment (figure \ref{fig:problem-scenarios}). 
The augmented information is the environment and occupant's ghost image and can have a significant impact when the primary visual system fails. 

\begin{figure*}[!t]
\centering
\includegraphics[width=2\columnwidth]{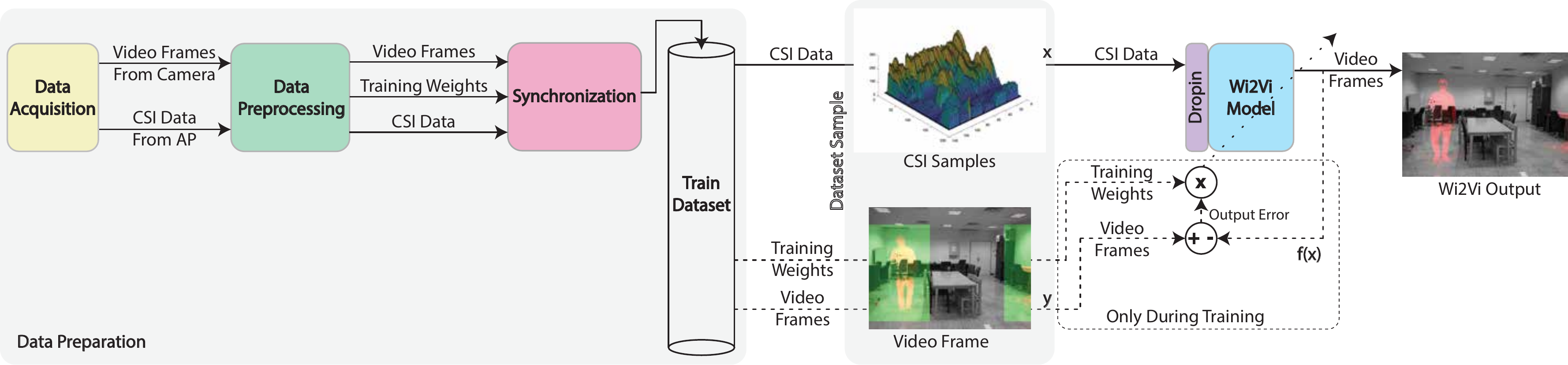}%
\caption{The Wi2Vi model dataflow}
\label{fig:data-processing-stages}
\end{figure*}
 

As it is depicted in figure \ref{fig:Normal-Scenario} and \ref{fig:Critical-Scenario}, this practical problem has two phases. In \textbf{normal conditions}, the Wi2Vi system trains and updates itself using the available images from the surveillance system's camera. As depicted in figure \ref{fig:Critical-Scenario} in \textbf{critical conditions} such as the time the room is filled with smoke or fog, or when the camera's output image quality is compromised, the conventional surveillance system fails to deliver acceptable outputs. In these situations, the previously trained Wi2Vi system generates video frames as images of humans inside the environment. In this paper, the camera and AP are assumed to be static, and their positions are assumed to be identical during both phases. Compared to the camera output, the quality and clarity of the generated video from the CSI data is limited, but in critical situations, every bit of information counts. 

Figure \ref{fig:data-processing-stages} depicts the steps in Wi2Vi model development. As we will discuss in the following sections and in order to develop the model, CSI samples and video frames are acquired and preprocessed. The polished data samples are then synchronized and stored in a training dataset. Wi2Vi uses CSI samples to generate video frames and uses captured video frames to estimates it's error and optimize it's performance.

The system uses CSI signals to generate video frames; see right side of figure \ref{fig:data-processing-stages}. If we denote the CSI data with $x$ and video frames as $y$, we can express our Wi2Vi model as
\begin{equation}
	y=f(x) \label{eq:sim-basic-model-1}.
\end{equation}

The dataset sample pairs $(x,y)$ are CSI samples and video frames required for Wi2Vi training.
During the training phase, the model denoted as $f$ is optimized using $x$ and $y$ pairs stored in the training dataset, which we denote as $D$. The solid lines on the right side of figure \ref{fig:data-processing-stages} denote this video frame generation flow.
The training phase solves the following optimization problem over the model parameters, $\theta_f$. 
\begin{equation}
    \forall    (x,y)\in D : \min_{\theta_f} {\mathcal{L}(y,f(x)) + \mathcal{R}(\theta_f)} \label{eq:model-opt-1}
\end{equation}

The model updates its parameters using the error between $f(x)$ and the paired video frame in dataset $y$. The dashed lines in figure \ref{fig:data-processing-stages} represent this model update flow.
The loss function, denoted as $\mathcal{L}$ in \eqref{eq:model-opt-1}, penalizes the error in the model output, and the regularizer $\mathcal{R}$, enforces our assumptions and knowledge on model parameters. Conventional choices for the loss function in image generation problems are the $L2$ and $L1$ norms on the output error. 
We chose the $L1$ distance between the actual video frame and what the Wi2Vi model predicted since it produces a model with sharper output images. 
Wi2Vi is a deep neural network, hence the stochastic optimization with Adam optimizer \cite{kingma2014adam} is our optimization technique of choice.
 
In the practical scenario depicted in figure \ref{fig:problem-scenarios}, the Wi2Vi system determines the relationship between CSI variations due to human presence in the room and its visual mapping to the camera output. The human body changes the reflection pattern of wireless signals according to the multipath signal propagation model \cite{tse2005fundamentals}. 
\begin{equation}
x_{\text{ch. output}}(t)=\sum_{i} \alpha(i) x_{\text{ch. input}}(t-\tau_i(t)) + n(t),\label{eq:multipath-chanel-model-1}
\end{equation}
where $x_{\text{ch. input}}(t)$ and $x_{\text{ch. output}}(t)$ are the channel input and output signals and $n(t)$ represents the noise in channel. Each reflection path $i$ face a different attenuation factor $\alpha(i)$ and has a delay $\tau_i(t)$ before reaching the receiving end.
Hence the CSI data varies according to the position of all objects in the environment including the human occupants. The Wi2Vi learns these variations using its \textit{CSI Encoder} part, translates this information to visual information, and creates the video frames using its \textit{Frame decoder} part (figure \ref{fig:CSI-env-Im-model}).


\begin{figure*}[!t]
\centering
\subfloat[Room view]{\includegraphics[width=1.44\columnwidth]{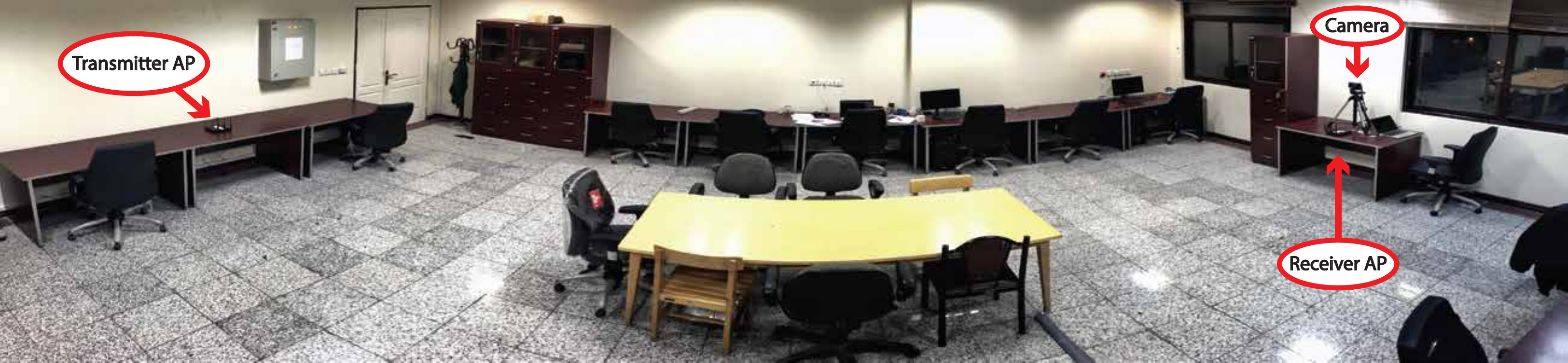}%
\label{fig:eval-setup-lab}}
\hfil
\subfloat[Transmitter point of view]{\includegraphics[width=.51\columnwidth]{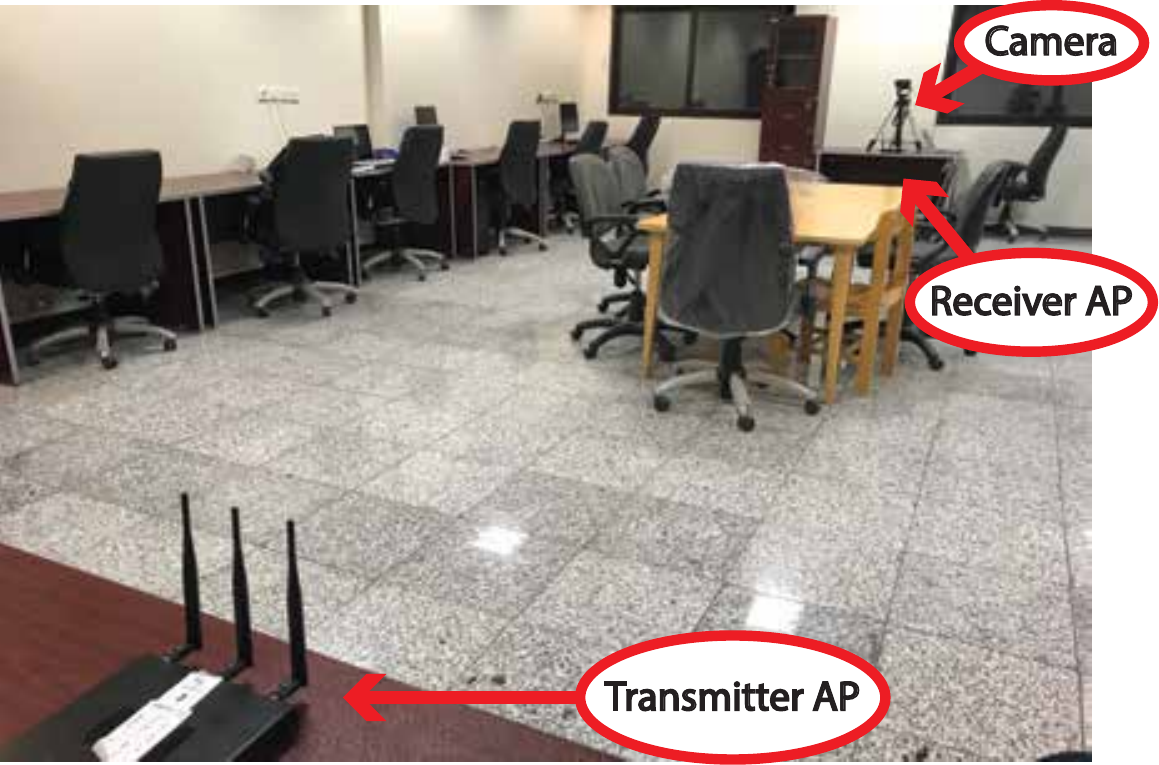}%
\label{fig:eval-setup-lab-side}}
\caption{The Wi2Vi evaluation setup for CSI and video acquisition}
\label{fig:eval-setup-lab-full}
\end{figure*}

\subsection{Data Preparation}
\subsubsection{Data Acquisition\label{sec:data-acq}}

The Wi2Vi requires a transmitter-receiver pair of APs and a camera. Figure \ref{fig:eval-setup-lab-full} shows our setup for signal acquisition. In this setup, the receiver AP and the camera were placed on one side of the room (right side of figure \ref{fig:eval-setup-lab}). The transmitter AP was placed on the table at the other side of the room (left side of figure \ref{fig:eval-setup-lab}). We provide more details on this setup configuration in section \ref{sec:eval-setup}.


	\textbf{{A. Camera's video frames:}}~
	 While the APs capture CSI data in both the training and execution phase, the camera only captures video frames during the model training phase. Thus the surveillance camera we use for capturing video frames may be removed after this data acquisition phase. The "dropin" method we discuss in section \ref{sec:dropin} will considerably reduce the signal acquisition duration, i.e., camera presence, by enabling the CSI data augmentation. These improve the privacy concerns of persons regarding the presence of a camera in the room. 

	 \textbf{{B. CSI measurements:}}~
The CSI is an estimate of the channel frequency response (CFR), and the receiver AP measures it for every packet reaching its physical layer. The receiver estimates the channel response in the frequency domain for all transmit-receive antenna pairs via the model defined in \eqref{eq:chanel-model-1}.
\begin{equation}
Y_{i}=H_{i} ~ X_{i} + N_i \label{eq:chanel-model-1}
\end{equation}

In this equation, the $i$ represents the frequency (subcarrier index) at which we are estimating the channel response. $H_{i}$ is a matrix of size $N_{TX}\times N_{RX}$ and includes all transmit-receive antenna pairs. $N_i$ is the noise part on each receiver antenna $Y_{i}$ and has the same dimension, $N_{RX}$. $X_{i}$ is the preamble signal of dimension $N_{TX}$ from the transmit antennas. The preamble length varies according to the wireless link speed but has a known structure to the receiver. For a fixed subcarrier index $i$, the ratio of the known $x_{t}$ and the received $y_{r}$, estimates the channel response for subcarrier $i$, between transmit antenna $t$ and receive antenna $r$.
\begin{equation}
h_{r,t} = \frac{y_r}{x_t}
\end{equation}

Since $h_{r,t}$ is complex-valued, we can represent it in amplitude and phase representation as \eqref{eq:h-com-amp}.
\begin{equation}
h_{r,t} = |h_{r,t}|e^{j \angle h_{r,t}} \label{eq:h-com-amp}
\end{equation}

We have the channel matrix estimate for each subcarrier from the OFDM subcarriers. Hence our channel estimate is a complex-valued 3D tensor with dimensions of subcarrier index, transmit and receive antenna numbers. 
\begin{figure}[!t]
\centering
\includegraphics[width=.5\columnwidth]{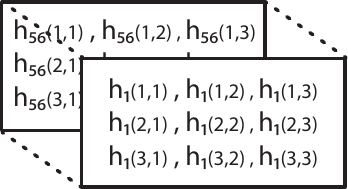}
\caption{A single CSI sample structure $S_i$, 3D tensor of size $(56\times 3\times 3)$}
\label{fig:CSI-H-matrix}
\end{figure}

For a typical 802.11n setup with MIMO capability, each AP has three antennas, and in $20MHz$ bandwidth mode, there are 56 subcarriers, which results in a CSI tensor of size $56\times 3\times 3$. Figure \ref{fig:CSI-H-matrix} depicts a CSI tensor for this case and we will refer to this CSI sample structure as $S_i$ from now on. Unfortunately, the CSI values reported by the COTS APs are not the exact CFR \cite{keerativoranan2018mitigation} as they experience different noise, interference sources, and signal degradations. Moreover, wireless network interface cards, e.g., Intel-5300, normalize the CSI values according to the RSSI value before reporting them. Nevertheless, the CSI data reported by the WiFi devices is related to the CFR, and WS models can extract required information for their task.
\begin{figure}[!t]
\centering
\includegraphics[width=1\columnwidth]{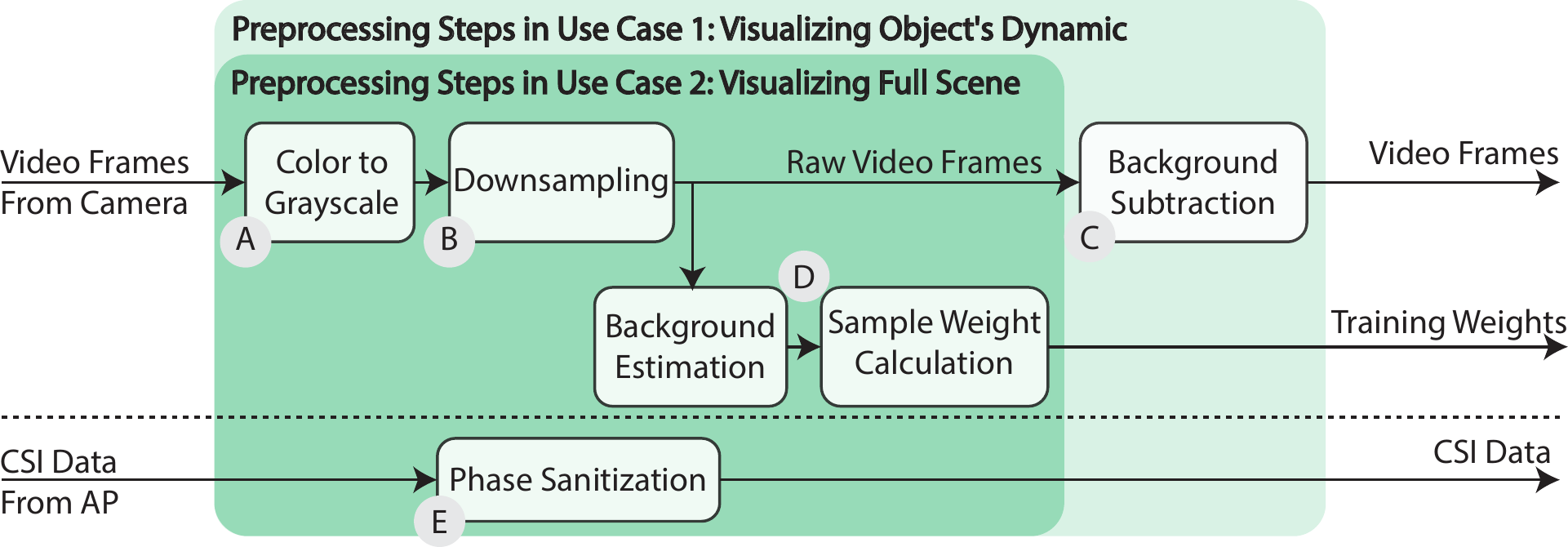}
\caption{Data preprocessing steps for the two evaluations}
\label{fig:Preprocessing-spets}
\end{figure}

\begin{figure*}[!t]
\centering
\includegraphics[width=2\columnwidth]{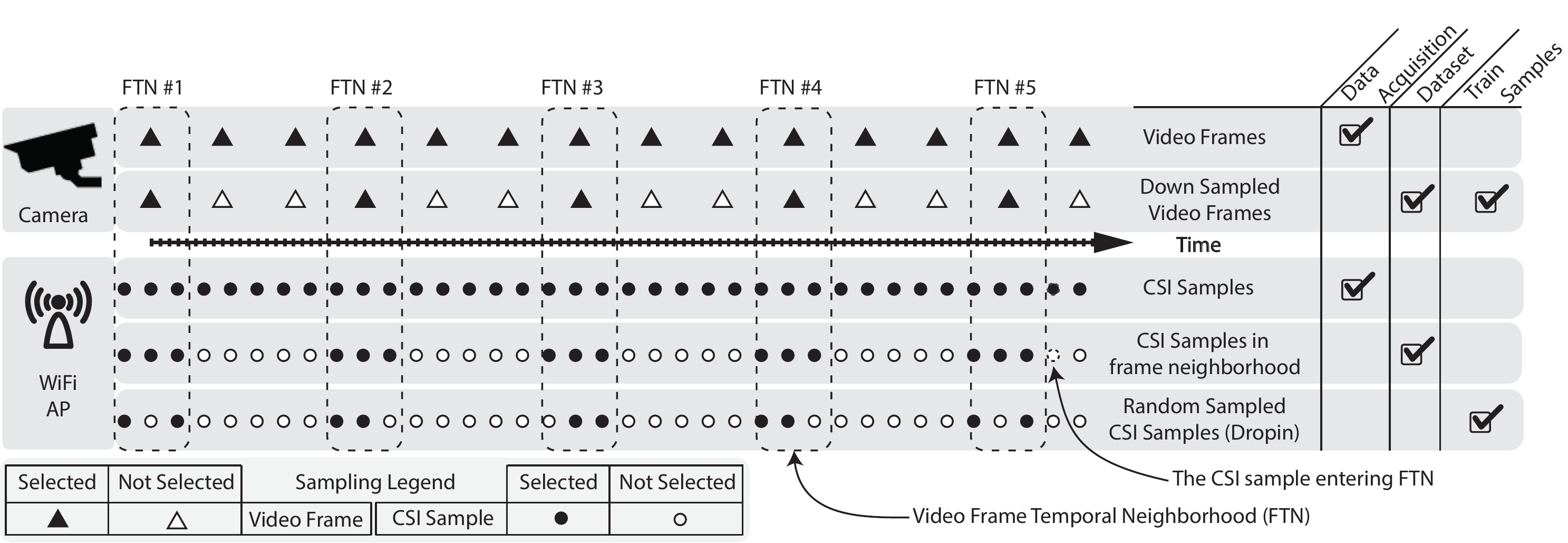}%
\caption{Video and CSI streams in data acquisition and dataset creation}
\label{fig:Dataset-general}
\end{figure*}

\subsubsection{Data preprocessing\label{sec:preprocessing} }
The preprocessing stage in figure \ref{fig:data-processing-stages} process CSI and video data and prepares the training dataset from the acquired data. Since we will present two use cases for the Wi2Vi model, two related preprocessing modes are preformed. Figure \ref{fig:Preprocessing-spets} depicts the overlapping nature of these preprocessing modes. We will now explain these steps according to their data scope.


	\textbf{{A. The video frames:}}~
The video frames are preprocessed before being synchronized to be stored in the training dataset. The preprocessing steps for the video frames are depicted in figure \ref{fig:Preprocessing-spets} with the details as below:

\textbf{\textit{A.1 Color to grayscale conversion:}}
At first, we transfer the acquired video frames from the color images to grayscale images, part A in figure \ref{fig:Preprocessing-spets}.

\textbf{\textit{A.2 Downsampling and resizing:}}
The indoor scene changes mostly happen at a low rate. Hence, the video samples from the camera are very correlated. In order to obtain a diverse and compact dataset, we reduce this redundancy by downsampling the video frames in time. 
The triangles in figure \ref{fig:Dataset-general} (above the gridded timeline in the middle) represent video frames captured by the camera and the black triangles in the second row of figure \ref{fig:Dataset-general} represent the video frames downsampled in time by three.

To match the captured video frames to our model's output dimensions, another function the downsampling does is resizing the grayscale frames, i.e., downsampling in space, step B in figure  \ref{fig:Preprocessing-spets}.

\textbf{\textit{A.3 Background removed images:}}
For the use cases where we are interested in visualizing the object's dynamics, we can focus the model's attention on the dynamics in the environment, i.e., humans in the scene like the practical scenario we presented in figure \ref{fig:problem-scenarios}, using the basic frame differencing approach, step C in figure \ref{fig:Preprocessing-spets}.

 Since we have assumed a static camera and AP setup, the background is static and may be estimated once using scenes with no dynamic objects, e.g., human subjects. We do not assume anything on the dynamic objects, and the Wi2Vi model is able to understand the dynamic effects of any object it senses during the training phase.

\textbf{\textit{A.4 Visual sample weighting\label{sec:samp-wighting}:}}
Another obstacle our video generating model faces is the simple fact that the objects closer to the camera produce a larger visual footprint due to the camera's perspective. If we use preprocessed video frames naively, the model biases to the samples relating to objects nearer to the camera. This bias prevents the effective modeling of CSI to video frames by Wi2Vi. Hence, to compensate for this bias, we add a regularizing weight term to each video frame used for training, step D in figure~\ref{fig:Preprocessing-spets}. 

The visual sample weight we use for each frame consists of multiple rectangular masks (the green rectangles in the dataset sample in the middle of figure \ref{fig:data-processing-stages}) and a weight term proportional to objects spread in the visual field (defined as $O({frame}_i)$).

\begin{equation}
	w_{frame_i}= 1 - O({frame}_i)
\end{equation}

We define the objects spread in the visual field as the number of pixels in the background removed video frame, $B({frame}_i)$.
\begin{equation}
	O(frame_i) = \frac{\#(B(frame_i)>0)} {P} \label{eq:visul-sprd}
\end{equation}

In \eqref{eq:visul-sprd}, $\#$ represents the number of elements, and $P$ is the total number of pixels in the video frame.

~

{\textbf{B. The CSI data:}}~
In order to use raw CSI measurements that are collected on the physical layer of the receiver AP, we have to adjust two issues: \textit{a) Variable time lag between CSI measurements}, \textit{b) Known phase uncertainties}. We deal with the latter in the data preprocessing and leave the former to section \ref{sec:dropin}.

\textbf{\textit{B.1 Removing variable time lags:}}
The transmitter transmits a preamble at the beginning of each packet, and thus the receiver obtains a channel estimate each time it receives a packet. In WS, the application on the transmitting AP periodically transmits packets to the receiver and stops for a fixed period between each transmission. This procedure results in channel estimates at nearly a constant speed, but the receiver drops packets with physical layer error. Hence, the exact time lag between CSI samples is variable. The dropin approach presented in section \ref{sec:dropin} solves not only this issue but also augments the dataset.

\textbf{\textit{B.2 Phase sanitization:}}
As it is widely known, raw CSI samples need phase sanitization before any use \cite{qian2018enabling,keerativoranan2018mitigation}. In our data preprocessing, step E in figure \ref{fig:Preprocessing-spets}, we sanitize the the phase of consecutive CSI samples using the linear sanitization scheme suggested in \cite{sen2012you}. CSI phase sanitization derives $\alpha_1$ and $\alpha_0$ using \eqref{eq:phase-san-1}. In \eqref{eq:phase-san-1}, $\phi_f$ is the phase of CSI measurement at subcarrier index $f$ and $f \in \{1, ... ,F\}$.
%
%
\begin{equation}
\label{eq:phase-san-1}
  \begin{gathered}
  \begin{aligned}
    \alpha_1&=\frac{\phi_F - \phi_1}{2\pi F}, ~~~~~~~~~~~~ 
	\alpha_0&=\frac{1}{F}\sum_{1\leq f\leq F} {\phi_f}
	\end{aligned} 
  \end{gathered}
\end{equation}

To obtain the sanitized phase values, \eqref{eq:phase-san-2} uses these two values.

\begin{equation}
	\hat{\phi_f} = \phi_f - (\alpha_1~f +\alpha_0)\label{eq:phase-san-2}
\end{equation}

~

\subsubsection{Synchronization of Video frames and CSI samples\label{sec:data-synchronization} }
The synchronization of the two data streams used by Wi2Vi has a critical impact on model performance. The asynchronous data stream degrades the model ability in demystifying the mapping from the CSI samples to the video frames. 

In the channel probing method discussed earlier, the transmitter's actual time of packet transmission may vary due to the traffic load in its network stack, and on the receiving end, random packet detection time results in considerable time jitter between the CSI samples. 
This time lag varies the time between CSI samples and degrades the periodicity of received CSI samples and their accordance with video frames. Hence, relying on counting the CSI samples index in time for synchronizing the CSI and video stream is incorrect.
To solve this problem, using the available time trace of CSI and video frame samples, we synchronize the downsampled (in time) video frames with CSI samples by selecting n-nearest CSI samples in time to the frame's capturing time, and call them the video frame's temporal neighborhood (FTN). 

The time difference between the video frame and CSI samples in each FTN is not fixed. For example, in figure \ref{fig:Dataset-general}, for each three video frames, eight CSI samples are captured. Since eight is not devisable by three, after four synchronized FTNs, and in the fifth FTN, a CSI sample is lagged back nearly enough to let another sample get near to the temporal neighborhood. In figure \ref{fig:Dataset-general}, the gray sample adjacent to the fifth temporal neighborhood is entering the fifth FTN. Hence, in the sixth neighborhood (which is not shown in figure \ref{fig:Dataset-general}), a CSI sample needs to be shifted when we select the sixth FTN. 

\subsubsection{The Dataset samples}
Following the preprocessing and synchronization procedure discussed earlier, downsampled video frames with their corresponding phase sanitized CSI samples in FTN are stored as dataset samples. The right table in figure \ref{fig:Dataset-general} clarifies the samples acquired, stored, and used on each dataset's data query.
 
\subsection{The dropin layer\label{sec:dropin}}
%
 Even though we can capture several samples for training Wi2Vi, capturing more samples requires more time, memory, and technical effort. We reduce the duration data acquisition procedure by proposing a new approach to augment captured samples. The proposed "dropin" layer for this task highly improves the model robustness to the random time lag between CSI samples mentioned earlier.
 
The dropin adds randomness in the model's input and has similarities to the widely appreciated and utilized dropout technique \cite{srivastava2014dropout}. The dropin layer randomly selects some samples from the CSI samples in FTN and reduces the sample's dimension in this way. 
Each CSI sample in FTN is a tensor $S_i\in {\mathbb{C}}^{56\times  3 \times 3}$ (figure \ref{fig:CSI-H-matrix}). Hence we can represent the CSIs in FTN as $X$, by concatenating CSI samples,
\begin{equation}
	X=[S_0, S_1, \hdots	, S_n]^T,
\end{equation}
where $X\in {\mathbb{C}}^{n\times { \text{size(}}S_i{\text{)}} } $ and $(.)^T$ represents the matrix transpose operation. So, the model's input data tensor is represented as $X$, and we can now define the dropin layer denoted as $D_{\text{dropin}}$ by  

\begin{equation}
X'^{T}=X^{T} ~ D_{\text{dropin}}. \label{eq:dropin-rel-1}
\end{equation}

In this equation, $X'^{T}$ is the dropin layer's output.
For the case where we have $n$ time samples for each dataset sample, and we want to select $k$ one of them to be used as the input of the next layer, $D_{\text{dropin}}$ is a matrix of size $n\times k$ with one non-zero element in each column equal to one. The  $D_{\text{dropin}}$ matrix can be easily constructed by randomly selecting $k$ columns of the identity matrix of size $n$. Note that $D_{\text{dropin}}$ is varied randomly on every data use by the model.
%
The bottom row of circles in figure \ref{fig:Dataset-general} depicts the randomly selected CSI samples for each FTN using a $n=3$ and $k=2$ dropin layer. 
As an example, the $D_{\text{dropin}}$s corresponding to \emph{FTN\#1} and \emph{FTN\#2} in figure \ref{fig:Dataset-general} are 
\begin{equation}
D^{FTN\#1}_{\text{dropin}} = \left[ \begin{array}{cc}
1 & 0 \\
0 & 0 \\
0 & 1 \end{array} \right],
~~~~
D^{FTN\#2}_{\text{dropin}} = \left[ \begin{array}{cc}
1 & 0 \\
0 & 1 \\
0 & 0 \end{array} \right].
\label{eq:dropin-rel-2}
\end{equation}




\begin{figure*}[!t]
\centering
\includegraphics[width=2\columnwidth]{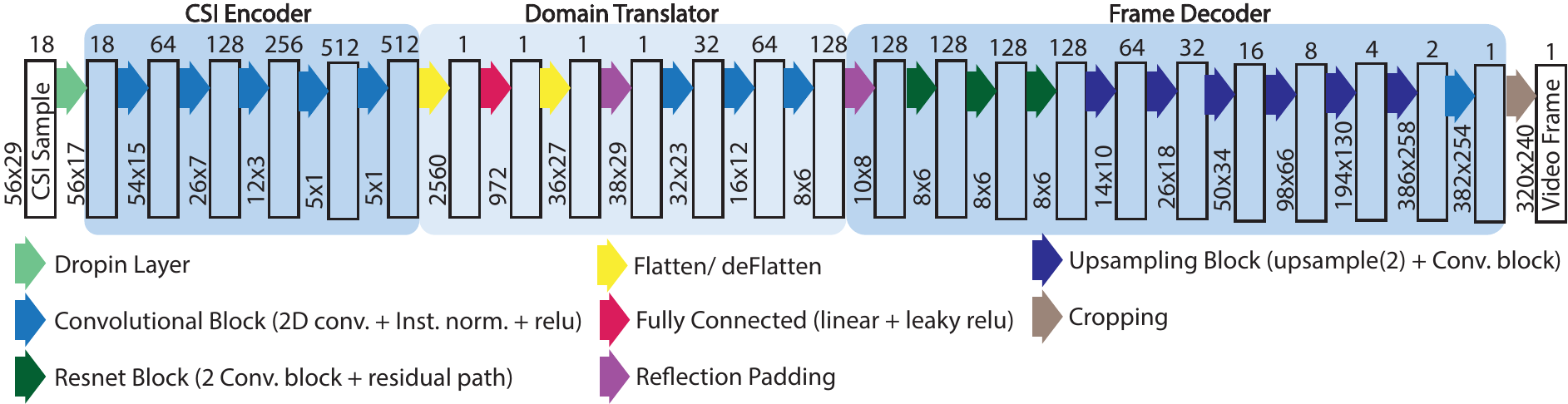}%
\caption{The Wi2Vi model structure}
\label{fig:Full-net-structure}
\end{figure*}

The dropin layer forces the model to learn the structures and forget the exact timing and orders, which means better generalization since the actual timings are not accurate. Since the randomness happens at every data fetch from the dataset, the model sees a new CSI sample at each epoch, even if we set the batch size to one. This may be interpreted as creating a new dataset with the size equal to the number of all combinations available in selecting randomly in the dropin layer.

\subsection{The Wi2Vi Model}
We now explain how the Wi2Vi model translates the CSI data to video frames. The proposed deep model includes multiple layers stacked for the image generation task. The intuition behind developing such a model is that the environment is the same, and the CSI readings and camera video frames are two types of measurements from the same underlying environment. Figure \ref{fig:CSI-env-Im-model} demonstrate this causal relationship and figure \ref{fig:Full-net-structure} shows the full network structure. 
The dropin layer defined in section \ref{sec:dropin} reduces the dataset samples dimension in time before it reaches the Wi2Vi model.
The Wi2Vi model includes three parts, which will be  explained in details below: \textit{CSI Encoder}, \textit{Domain Translator}, \textit{Frame decoder}.

\subsubsection{CSI encoder}
The \textit{CSI encoder} is responsible for getting the CSI samples in FTN and generating its representation in the CSI's latent space. This latent space should capture some features out of the CSI samples such that the rest of the Wi2Vi model can produce their corresponding video frame. To construct such an encoder, we have used a deep convolutional neural network. More specifically, the \textit{CSI encoder} consists of five 2D convolutional layers, with instance normalization and relu activation function.





\subsubsection{Domain translator}
The \textit{CSI encoder} generates a latent domain representation of the CSI measurements. The next task is to generate the corresponding video frame from this latent domain representation. However, applying the \textit{Frame decoder} directly on this latent domain representations does not lead to video frames with satisfying visual quality. It is mainly due to that, even though CSI measurements and video frames are acquired in the same environment, they are in very different domains.
The \textit{Domain Translator} block 
 uses the canonical fully connected layer, followed by three convolutional blocks for this task.
Domain translation starts with a fully connected (a linear and a leaky relu layer) that reduces the data dimensions by a factor of approximately three. This bottleneck structure burdens the model capacity, preventing it from overfitting to the train samples.

\subsubsection{Frame Decoder}

The last part of Wi2Vi generates video frames from the translated latent domain representation. Wi2Vi's \textit{Frame Decoder} network consists of two parts. Three ResNet \cite{he2016deep} parts firstly create the video frame in low dimensions and an upsampling part, which increases the output size, generating output video frames with the desired resolution. Each Resnet part has two convolutional blocks with a sum connection, and each upsampling block has an upsampling (by two) layer followed by a convolutional block which smoothes the upsampling results.



\section{Implementation and evaluation}
\subsection{Implementations}

We evaluate the proposed Wi2Vi model in two different use cases. 

In our first use case, we aim at visualizing object's dynamic in the environment.
In this use case, the model was trained on background removed video frames since we are not interested in the static background part of the video frames. 

In our second use case, we consider the visualizing the full scene and video frame generation. In this setting, the Wi2Vi model produces video frames with a static background and dynamic subject information.
The preprocessing of the video frames does not include background subtraction in this case (part E in figure \ref{fig:Preprocessing-spets}).


\subsection{Evaluation Setup \label{sec:eval-setup}}
For evaluation, we collected data in a 30 minutes long acquisition session. The room and the setup used for acquisition are depicted in figure \ref{fig:eval-setup-lab-full}. The room's dimensions were 9.5m x 6.5m, and as the APs were capturing CSI samples, a person was walking around the desk in the center of the room. 

\textbf{Camera's Video Frames:}
A Sony-alpha6000 camera captured the video data at 30 frames per second in VGA resolution (640 x 480) during the CSI signal acquisition. In the preprocessing phase, we reduced the frame size by two (320 x 240) and downsampled the frame rate by five, resulting in 6 video frames per second. 

The preprocessing steps introduced in section \ref{sec:preprocessing} for both use cases were carried out accordingly, and the data streams were synchronized (section \ref{sec:data-synchronization}).
%

\begin{figure}[!t]
\centering
\includegraphics[width=\columnwidth]{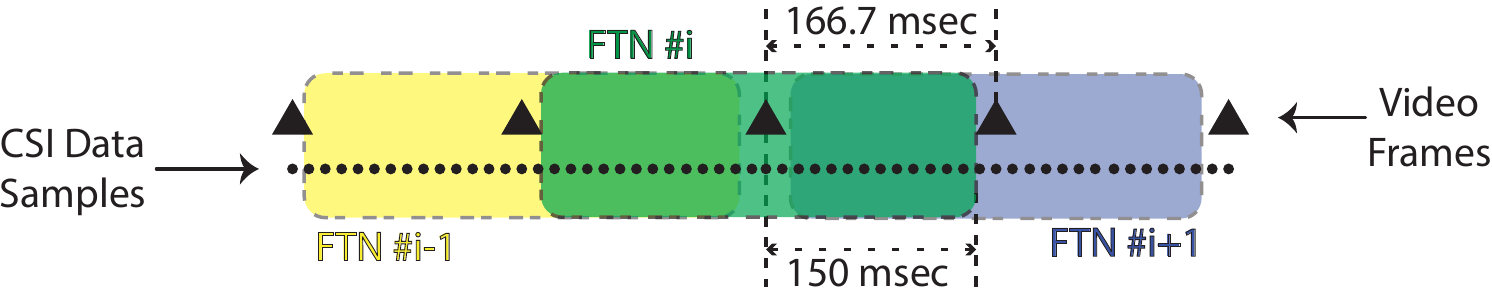}
\caption{Downsampled Video frames and CSI samples in temporal neighborhood structure}
\label{fig:my-data-temporal-neigh}
\end{figure}

\textbf{CSI Measurements:}
A transmit-receiver pair of Atheros APs with the modified driver \cite{Xie:2015:PPD:2789168.2790124-} was capturing the CSI data-trace during the test. We set the transmitter delay between packets to 10 msec resulting an approximately 100 CSI captures per second. APs work in 3x3 MIMO, 20 MHz bandwidth mode. Hence,  CSI samples returned by the driver have 56 subcarriers, three by three complex matrix (figure \ref{fig:CSI-H-matrix}). Each video frame stored in the dataset was associated with 29 CSI samples in FTN. This temporal neighborhood size results maximum CSI sample lag of approximately $\pm 150~msec$. Since we have downsampled video frames in time from 30 fps to 6 fps, the time difference between consecutive frames is $166.7~msec$, which is close to our maximum CSI sample lag, creating approximately $50\%$ overlap between FTNs (figure \ref{fig:my-data-temporal-neigh}).

\begin{figure*}[!t]
\centering
\includegraphics[width=2\columnwidth]{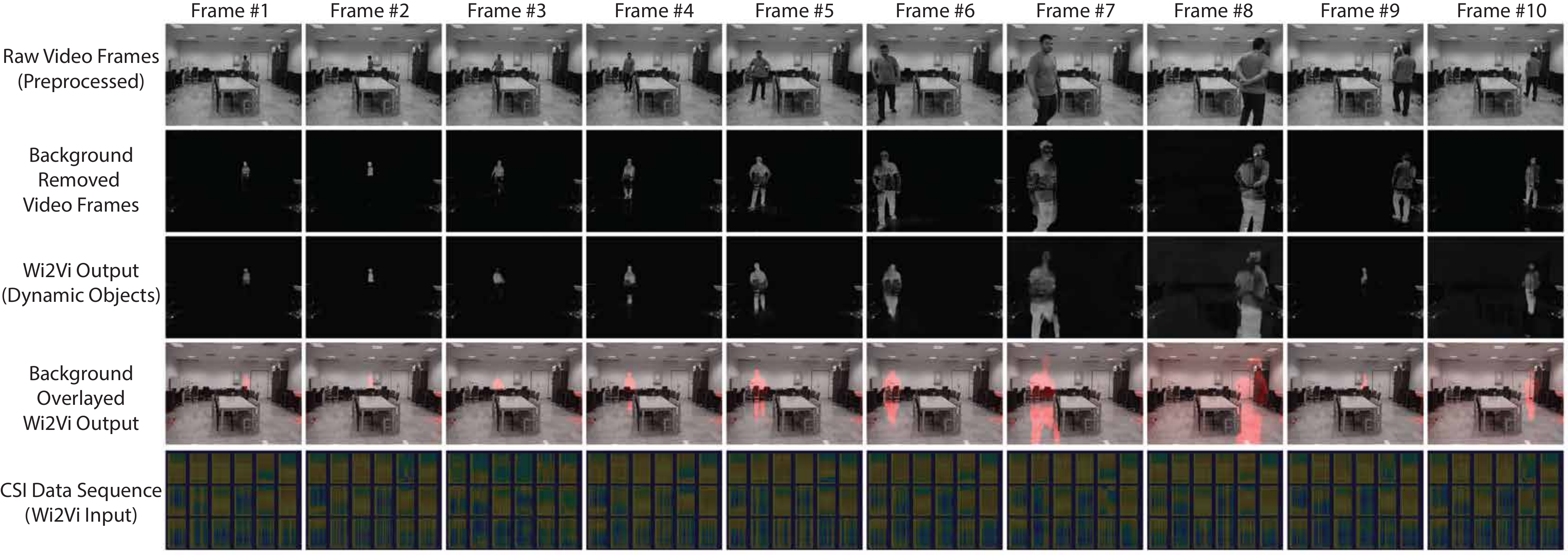}%
\caption{Wi2Vi Model evaluation results in case 1: dynamic Objects using background removed video frames}
\label{fig:eval-1-test-trace}
\end{figure*}

\begin{figure*}[!t]
\centering
\includegraphics[width=2\columnwidth]{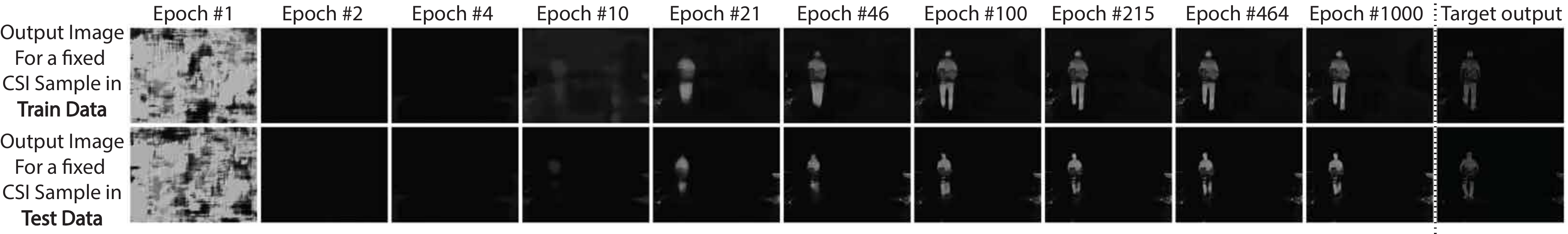}%
\caption{Wi2Vi Model convergence for a fixed sample from the train data and a similar data from the test data in the first evaluation}
\label{fig:eval-1-convergence}
\end{figure*}

\textbf{Environment Conditions:}
Throughout the signal acquisition, four persons were present in the room. One of these people was the main subject and randomly moved inside the room. We exclude samples obtained while other peoples have major movements since we focus on the single person scenario in this paper. However, small movements in other objects, e.g., the seats were also present in the data. 


\textbf{Train and Test Dataset:}
The resulting total number of samples in our dataset was $8,300$. We trained the model on the first $7,885$ samples ($95\%$) in the dataset, and the remaining last $415$ samples ($5\%$)  were used as the validation and test samples. For a better comparison of the results in the two parts, we conducted both evaluations with the same datasets. 

\textbf{Training Configurations:}
In training the model for both evaluations, the learning rate used in the model updates started from $0.002$ and was reduced $4.5\%$ every five epochs. The training procedure includes 1000 epochs with a batch size of $32$.

\section{Results}
We will now review the results obtained for the two use cases presented before. In the first use case, the model only visualizes the object's dynamics in the environment registered in background removed video frames, and in the second use case, the model targets visualizing the full scene (static and dynamic objects) in the scene using video frames. 

\begin{figure*}[!t]
\centering
\includegraphics[width=2\columnwidth]{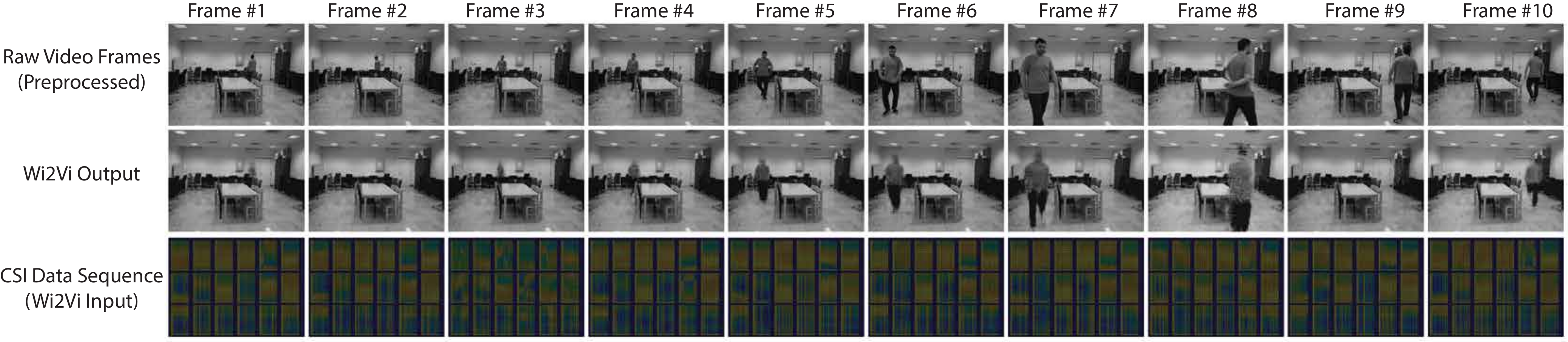}%
\caption{Wi2Vi Model evaluation results in case 2: static and dynamic Objects using raw video frames}
\label{fig:eval-2-test-trace}
\end{figure*}

\begin{figure*}[!t]
\centering
\includegraphics[width=2\columnwidth]{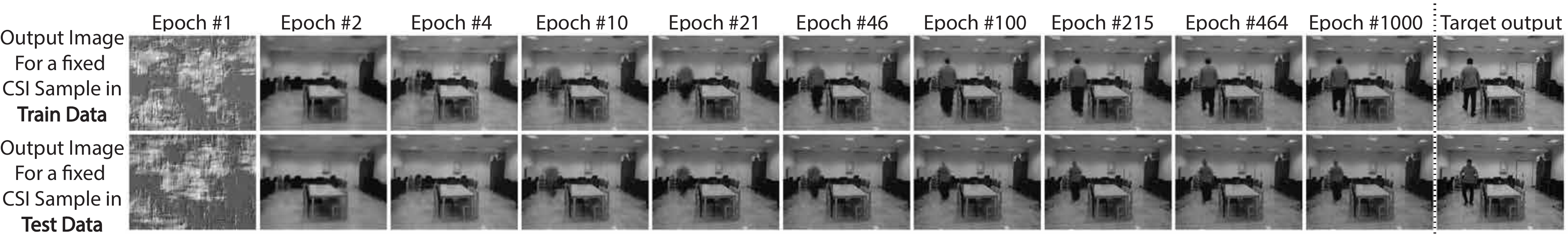}%
\caption{Wi2Vi Model convergence for a fixed sample from the train data and a similar data from the test data in the second evaluation}
\label{fig:eval-2-convergence}
\end{figure*}

\subsection{Use case I: Visualizing Object's Dynamic \label{sec:usecase-I} }
Training the Wi2Vi model with background removed video frames, Wi2Vi can produce similar video frames with the dynamic subject and no static background. Figure \ref{fig:eval-1-test-trace} depicts ten frames of the resulting video sequence in the evaluation of the Wi2Vi model over the test samples. Since we have separated train and test samples in time, the model had not seen the desired video output for these samples and is solely employing its knowledge acquired from the train samples.  In this figure, the first row depicts real frame samples. The second and the third row include the actual and model outputted background-subtracted video frames, respectively. The fourth row depicts the background imposed model outputs, where for better comparison with the first row, model output is overlayed by red color. The fifth row represents the Wi2Vi input data, i.e., the CSI data in FTN.

The generalization of the Wi2Vi model is acceptable in nine out of ten video frames depicted in figure \ref{fig:eval-1-test-trace}, and the model has produced unsatisfactory results in the ninth column. The visual quality of the produced video frames suffice basic surveillance needs, e.g., human presence detection and location estimation. 

Compared to the previous studies in WiFi imaging, e.g., \cite{zi2019wi}, the Wi2Vi significantly outperforms in output image (video frame) quality and execution requirements. Reference \cite{zi2019wi} produces images using five APs (one transmitter and four receivers). Moreover, they use a customized antenna configuration for the receiver APs. Since the Wi2Vi only uses two APs without any modifications, a direct fair comparison of \cite{zi2019wi} and Wi2Vi is not possible.


To observe the evolution of the Wi2Vi model during training, figure \ref{fig:eval-1-convergence} depicts the model output for a fixed CSI sample from the train and test datasets in different epochs. The epoch number for drawing the output image is chosen exponentially to depict the convergence more accurately. In figure \ref{fig:eval-1-convergence}, the column on the right includes the actual background removed video frames, i.e., target outputs. 
 
 As it can be seen in figure \ref{fig:eval-1-convergence}, after the first epoch, the model has realized the empty frame, i.e., no dynamic information in most area of the output video frame, and it has taken it twenty train epochs to understand the appropriate output according to the dynamics in the CSI data.
 
 The samples on the first and the second rows of figure \ref{fig:eval-1-convergence} are deliberately picked to have close visual domain representation. As it is noticeable, the Wi2Vi model has differentiated in small details of the subject's posture. This differentiation exposes the model generalization ability. 

\subsection{Use case II: Visualizing Full Scene}

For our second Wi2Vi model use case, we consider generating video frames with background information. The model has to learn the static background information along with the dynamic objects (the human moving in the environment). 

Figure \ref{fig:eval-2-test-trace} depicts ten frames that the Wi2Vi model has produced from the CSI test samples. The model's performance has decreased compared to the results presented for the previous use case since we have used the same model structure and training configuration for a more complicated task. 
The full modeling of objects is very challenging. However,
the model was able to differentiate between the static background and the dynamic human, results depicted in figure \ref{fig:eval-2-test-trace} point out the model's failure in creating video frames two and nine. 

Similar to the results presented for the use case one, figure \ref{fig:eval-2-convergence} portrays how the model's parameters convergence affects the produced output during the model training. As it can be observed, the model was able to determine static background details after the first epoch of training and has focused on the dynamics afterward.

\subsection{Sample Generated Videos}
The video frames we presented in figures \ref{fig:eval-1-test-trace} and \ref{fig:eval-2-test-trace} might not be clear enough. Therefore we have also provided the GIF files containing the original video frames and the generated video sequences for the two evaluation in
\href{https://github.com/mhkefayati/Wi2Vi.git}{$https://github.com/mhkefayati/Wi2Vi.git$}.

\section{Conclusion}
There exist a correlation between the effect of objects in the environment between different frequency spectrums. The Wi2Vi model proposed in this paper extracts this correlation between the CSI data in the WiFi 2.4 GHz band and the video frames in the visual spectrum. After a brief sample acquisition session, the Wi2Vi system produces video frames entirely using WiFi CSI. The Wi2Vi is robust to visual domain signal limitations, e.g., specific lighting conditions. Furthermore, Wi2Vi only uses video frames for the initial training phase, and the camera may be removed during system execution, improving privacy concerns in conventional video surveillance systems.

We have presented results for two use cases of the Wi2Vi system implemented using commercial off the shelf (COTS) Atheros WiFi chipsets. The results obtained on both use cases confirm the Wi2Vi ability to produce dynamic and static visual information using CSI measurements entirely. Aside from the unique network structure of the Wi2Vi model, the satisfying performance of the Wi2Vi model relies on our data preparation, and the novel "dropin" layer that virtually enhances the dataset size and improves the model's robustness to the existing jitter in CSI measurements in COTS devices.

\section*{Acknowledgment}
The authors would like to thank Mr. Ali Ghelmani and Mr. Ali Gharari for their contributions during data acquisition. We are also thankful to Futurebound Corporation for providing the Atheros WiFi APs for our CSI measurements.

\ifCLASSOPTIONcaptionsoff
  \newpage
\fi



\bibliographystyle{IEEEtran}
\bibliography{IEEEabrv,citations.bib}

\begin{thebibliography}{10}
\providecommand{\url}[1]{#1}
\csname url@samestyle\endcsname
\providecommand{\newblock}{\relax}
\providecommand{\bibinfo}[2]{#2}
\providecommand{\BIBentrySTDinterwordspacing}{\spaceskip=0pt\relax}
\providecommand{\BIBentryALTinterwordstretchfactor}{4}
\providecommand{\BIBentryALTinterwordspacing}{\spaceskip=\fontdimen2\font plus
\BIBentryALTinterwordstretchfactor\fontdimen3\font minus
  \fontdimen4\font\relax}
\providecommand{\BIBforeignlanguage}[2]{{%
\expandafter\ifx\csname l@#1\endcsname\relax
\typeout{** WARNING: IEEEtran.bst: No hyphenation pattern has been}%
\typeout{** loaded for the language `#1'. Using the pattern for}%
\typeout{** the default language instead.}%
\else
\language=\csname l@#1\endcsname
\fi
#2}}
\providecommand{\BIBdecl}{\relax}
\BIBdecl

\bibitem{anderson1993ray}
H.~R. Anderson, ``A ray-tracing propagation model for digital broadcast systems
  in urban areas,'' \emph{IEEE Transactions on Broadcasting}, vol.~39, no.~3,
  pp. 309--317, 1993.

\bibitem{glassner1989introduction}
A.~S. Glassner, \emph{An introduction to ray tracing}.\hskip 1em plus 0.5em
  minus 0.4em\relax Elsevier, 1989.

\bibitem{zi2019wi}
Y.~Zi, W.~Xi, L.~Zhu, F.~Yu, K.~Zhao, and Z.~Wang, ``Wi-fi imaging based
  segmentation and recognition of continuous activity,'' in \emph{International
  Conference on Collaborative Computing: Networking, Applications and
  Worksharing}.\hskip 1em plus 0.5em minus 0.4em\relax Springer, 2019, pp.
  623--641.

\bibitem{wang2019person}
F.~Wang, S.~Zhou, S.~Panev, J.~Han, and D.~Huang, ``Person-in-wifi:
  Fine-grained person perception using wifi,'' \emph{arXiv preprint
  arXiv:1904.00276}, 2019.

\bibitem{wang2019can}
F.~Wang, S.~Panev, Z.~Dai, J.~Han, and D.~Huang, ``Can wifi estimate person
  pose?'' \emph{arXiv preprint arXiv:1904.00277}, 2019.

\bibitem{wang2018csi}
F.~Wang, J.~Han, S.~Zhang, X.~He, and D.~Huang, ``Csi-net: Unified human body
  characterization and pose recognition,'' \emph{arXiv: 1810.03064}, 2018.

\bibitem{srivastava2014dropout}
N.~Srivastava, G.~Hinton, A.~Krizhevsky, I.~Sutskever, and R.~Salakhutdinov,
  ``Dropout: a simple way to prevent neural networks from overfitting,''
  \emph{The journal of machine learning research}, vol.~15, no.~1, pp.
  1929--1958, 2014.

\bibitem{Halperin_csitool}
D.~Halperin, W.~Hu, A.~Sheth, and D.~Wetherall, ``Tool release: Gathering
  802.11n traces with channel state information,'' \emph{ACM SIGCOMM CCR},
  vol.~41, no.~1, p.~53, Jan. 2011.

\bibitem{Xie:2015:PPD:2789168.2790124-}
Y.~Xie, Z.~Li, and M.~Li, ``Precise power delay profiling with commodity
  wifi,'' in \emph{Proceedings of the 21st Annual International Conference on
  Mobile Computing and Networking}, ser. MobiCom '15.\hskip 1em plus 0.5em
  minus 0.4em\relax New York, NY, USA: ACM, 2015, pp. 53--64.

\bibitem{ma2019wifi}
Y.~Ma, G.~Zhou, and S.~Wang, ``Wifi sensing with channel state information: A
  survey,'' \emph{ACM Computing Surveys (CSUR)}, vol.~52, no.~3, p.~46, 2019.

\bibitem{al2019channel}
M.~A. Al-qaness, M.~Abd~Elaziz, S.~Kim, A.~A. Ewees, A.~A. Abbasi, Y.~A. Alhaj,
  and A.~Hawbani, ``Channel state information from pure communication to sense
  and track human motion: A survey,'' \emph{Sensors}, vol.~19, no.~15, p. 3329,
  2019.

\bibitem{yousefi2017survey}
S.~Yousefi, H.~Narui, S.~Dayal, S.~Ermon, and S.~Valaee, ``A survey on behavior
  recognition using wifi channel state information,'' \emph{IEEE Communications
  Magazine}, vol.~55, no.~10, pp. 98--104, 2017.

\bibitem{jiang2018smart}
H.~Jiang, C.~Cai, X.~Ma, Y.~Yang, and J.~Liu, ``Smart home based on wifi
  sensing: A survey,'' \emph{IEEE Access}, vol.~6, pp. 13\,317--13\,325, 2018.

\bibitem{wang2019survey}
Z.~Wang, K.~Jiang, Y.~Hou, W.~Dou, C.~Zhang, Z.~Huang, and Y.~Guo, ``A survey
  on human behavior recognition using channel state information,'' \emph{IEEE
  Access}, vol.~7, pp. 155\,986--156\,024, 2019.

\bibitem{zhang2019comprehensive}
H.-B. Zhang, Y.-X. Zhang, B.~Zhong, Q.~Lei, L.~Yang, J.-X. Du, and D.-S. Chen,
  ``A comprehensive survey of vision-based human action recognition methods,''
  \emph{Sensors}, vol.~19, no.~5, p. 1005, 2019.

\bibitem{huang2014feasibility}
D.~Huang, R.~Nandakumar, and S.~Gollakota, ``Feasibility and limits of wi-fi
  imaging,'' in \emph{Proceedings of the 12th ACM Conference on Embedded
  Network Sensor Systems}.\hskip 1em plus 0.5em minus 0.4em\relax ACM, 2014,
  pp. 266--279.

\bibitem{zhou2015sensorless}
Z.~Zhou, C.~Wu, Z.~Yang, and Y.~Liu, ``Sensorless sensing with wifi,''
  \emph{Tsinghua Science and Technology}, vol.~20, no.~1, pp. 1--6, 2015.

\bibitem{adib2013see}
F.~Adib and D.~Katabi, \emph{See through walls with WiFi!}\hskip 1em plus 0.5em
  minus 0.4em\relax ACM, 2013, vol.~43, no.~4.

\bibitem{kotaru2015spotfi}
M.~Kotaru, K.~Joshi, D.~Bharadia, and S.~Katti, ``Spotfi: Decimeter level
  localization using wifi,'' in \emph{ACM SIGCOMM computer communication
  review}, vol.~45, no.~4.\hskip 1em plus 0.5em minus 0.4em\relax ACM, 2015,
  pp. 269--282.

\bibitem{holl2017holography}
P.~M. Holl and F.~Reinhard, ``Holography of wi-fi radiation,'' \emph{Physical
  review letters}, vol. 118, no.~18, p. 183901, 2017.

\bibitem{luo2018wi}
R.~Luo, Z.~Zhang, X.~Wang, and Z.~Lin, ``Wi-fi based device-free microwave
  ghost imaging indoor surveillance system,'' in \emph{2018 28th International
  Telecommunication Networks and Applications Conference (ITNAC)}.\hskip 1em
  plus 0.5em minus 0.4em\relax IEEE, 2018, pp. 1--6.

\bibitem{wang2016microwave}
X.~Wang and Z.~Lin, ``Microwave surveillance based on ghost imaging and
  distributed antennas,'' \emph{IEEE Antennas and Wireless Propagation
  Letters}, vol.~15, pp. 1831--1834, 2016.

\bibitem{zhang2019feasibility}
D.~Zhang, J.~Wang, J.~Jang, J.~Zhang, and S.~Kumar, ``On the feasibility of
  wi-fi based material sensing,'' in \emph{The 25th Annual International
  Conference on Mobile Computing and Networking}.\hskip 1em plus 0.5em minus
  0.4em\relax ACM, 2019, p.~41.

\bibitem{kingma2014adam}
D.~P. Kingma and J.~Ba, ``Adam: A method for stochastic optimization,''
  \emph{arXiv preprint arXiv:1412.6980}, 2014.

\bibitem{tse2005fundamentals}
D.~Tse and P.~Viswanath, \emph{Fundamentals of wireless communication}.\hskip
  1em plus 0.5em minus 0.4em\relax Cambridge university press, 2005.

\bibitem{keerativoranan2018mitigation}
N.~Keerativoranan, A.~Haniz, K.~Saito, and J.-i. Takada, ``Mitigation of csi
  temporal phase rotation with b2b calibration method for fine-grained motion
  detection analysis on commodity wi-fi devices,'' \emph{Sensors}, vol.~18,
  no.~11, p. 3795, 2018.

\bibitem{qian2018enabling}
K.~Qian, C.~Wu, Z.~Yang, Y.~Liu, F.~He, and T.~Xing, ``Enabling contactless
  detection of moving humans with dynamic speeds using csi,'' \emph{ACM
  Transactions on Embedded Computing Systems (TECS)}, vol.~17, no.~2, p.~52,
  2018.

\bibitem{sen2012you}
S.~Sen, B.~Radunovic, R.~R. Choudhury, and T.~Minka, ``You are facing the mona
  lisa: Spot localization using phy layer information,'' in \emph{Proceedings
  of the 10th international conference on Mobile systems, applications, and
  services}.\hskip 1em plus 0.5em minus 0.4em\relax ACM, 2012, pp. 183--196.

\bibitem{he2016deep}
K.~He, X.~Zhang, S.~Ren, and J.~Sun, ``Deep residual learning for image
  recognition,'' in \emph{Proceedings of the IEEE conference on computer vision
  and pattern recognition}, 2016, pp. 770--778.

\end{thebibliography}
\end{document}